\documentclass{article}
\usepackage{spconf,amsmath,graphicx}
\usepackage{booktabs}
\usepackage{makecell, multirow}
\usepackage{xcolor}

\title{Does Video Summarization require Videos? \\Quantifying the Effectiveness of Language in Video Summarization}

\name{Yoonsoo Nam*, Adam Lehavi, Daniel Yang, Digbalay Bose, Swabha Swayamdipta, Shrikanth Narayanan \thanks{* Corresponding Author. Email: yoonsoon@usc.edu}}
\address{University of Southern California}
\begin{document}
%
\maketitle
\begin{abstract}
Video summarization remains a huge challenge in computer vision due to the size of the input videos to be summarized.
We propose an efficient, language-only video summarizer that achieves competitive accuracy with high data efficiency. 
Using only textual captions obtained via a zero-shot approach, we train a language transformer model and forego image representations. 
This method allows us to perform filtration amongst the representative text vectors and condense the sequence. 
With our approach, we gain explainability with natural language that comes easily for human interpretation and textual summaries of the videos. 
An ablation study that focuses on modality and data compression shows that leveraging text modality only effectively reduces input data processing while retaining comparable results. 
\end{abstract}
\section{Introduction}
\label{sec:intro}
Automatic video summarization techniques have gained interest in recent years. Large volumes of video data are created and distributed on the Internet, requiring improved data management and access systems. Because an efficient video summarization technique can alleviate the challenges caused by big video data, the topic of automatic video summarization has been widely studied.

The task of automated video summarization has been primarily cast as a vision task. Recent related literature formulates video summarization as an image sequence-to-sequence classification problem which has inputs of image vectors and outputs of binary keyframe classification \cite{vslstm,mahasseni2017unsupervised,rochan2018video,zhu2020dsnet,feng2018extractive, narasimhan2021clip, apostolidis2021combining}. However, such a formulation poses issues. A major challenge is reducing the input size by appropriately addressing the redundancies while preserving novelty in the image sequences in a video. In prior work, typically this is accomplished by ad hoc subsampling the video at a rate that is smaller than the original video frame rate.  However, using this method does not ensure that the information from the original sequence of images with a higher frame rate will be preserved in the smaller subsample. The other challenge while deriving the neural representations using downstream is the loss of interpretability. 
For example, when using GoogLeNet \cite{szegedy2015going} to embed the image vectors, the model passes the original image signal RGB matrices 
through 22 layers to represent the image with a significantly smaller dimension of 1024. The complex mapping between the resultant vector and images makes it difficult to interpret the input used for the models.

Instead of posing video summarization as purely a vision modality problem, in this paper, we propose Language Modal Video Summarizer (LMVS), as a novel way of interpreting video frames through the lens of text. Firstly, by employing sentence embeddings instead of image embeddings, our model facilitates the filtration and condensing of image sequence representations, thereby enhancing data efficiency. This sequence compression further enables the inference of extended sequences. Second, natural language is conducive to easier human interpretation. In our model, we tokenize and embed each sentence at the word level. In contrast to the image embeddings, the embedded tokens directly correspond to segments of text. This direct correlation helps us easily decipher what the model sees as inputs. Lastly, by exploiting the language modality, we can create an extractive text summary of the video using captions of selected frames. The text summary offers a textual overview of the video and elucidates the model's choice for enhanced explainability. 
\begin{figure*}[t]
  \centering
  \includegraphics[width=\linewidth, height=6.5cm]{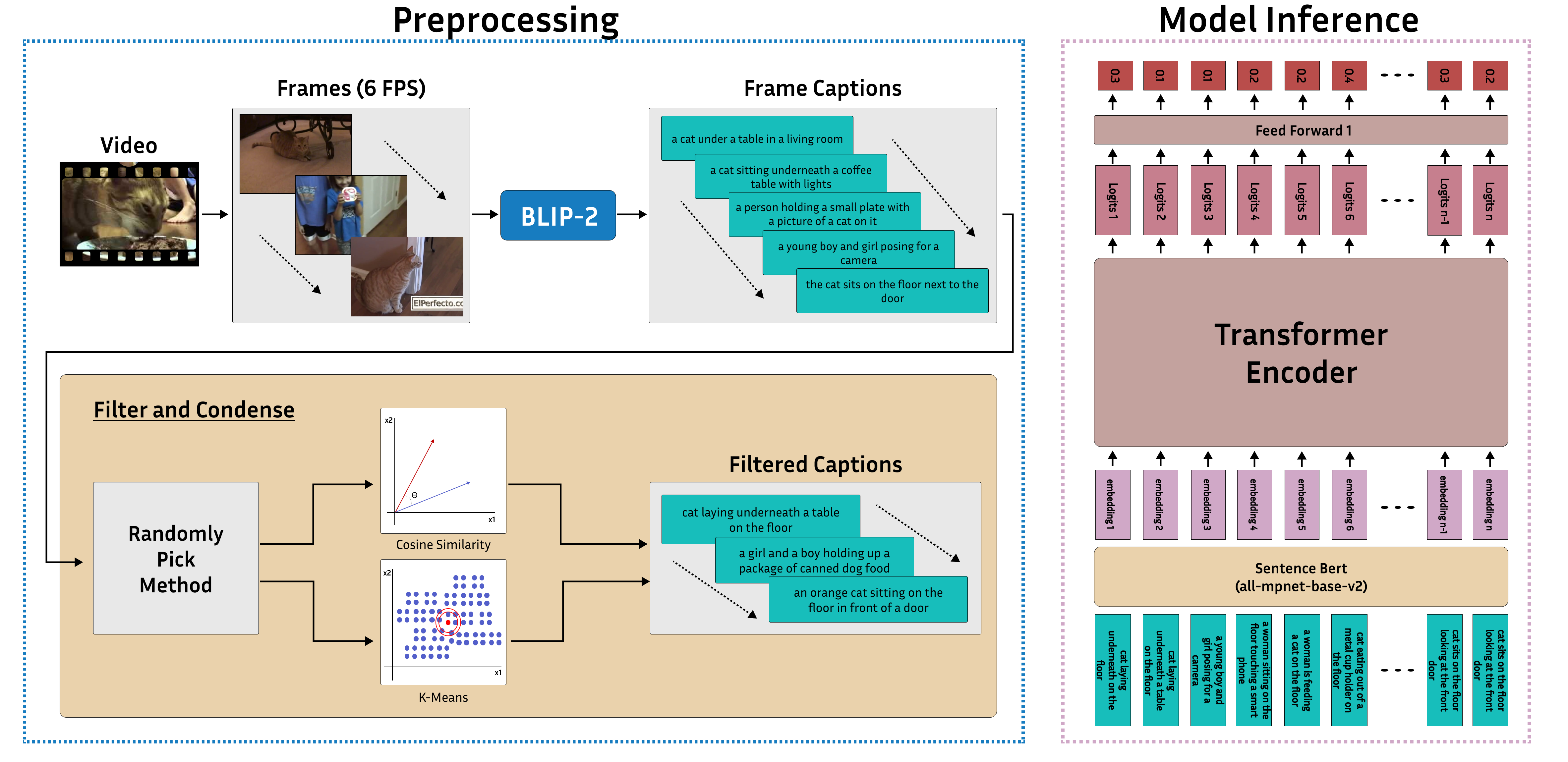}
  \caption{LMVS Model Pipeline. During preprocessing, we sample video at 6 fps, generate 3 captions per frame using BLIP-2, and filter and condense to 1 sentence per second. The filtered sentences are used to predict the frame-level importance score.}
  \label{fig:fullmodel}
\end{figure*}
\section{Language-Only Video Summarization}
\label{sec:methodology}
Figure \ref{fig:fullmodel} illustrates our approach comprising two stages: preprocessing and model inference. 
Preprocessing has three steps: (1) Videos to Frame Images; (2) Frame Images to Sentences; (3) Filter and Condense Sentences. 
The model inference also comprises three steps: (1) converting sentences to embeddings; (2) generating frame level importance scores; (3) generating a text summary. 
We uniformly sample the original video at the rate of 6 fps (frames per second). 
We use BLIP-2 \cite{li2023blip} to extract sentences from the frames. 
Then, we use a combination of cosine similarity and k-means \cite{kmeans} with k = 2 to filter the sentences generated for each second of the video. 
We embed the filtered sentences using SentenceBERT (SBERT) \cite{sbert}. 
Finally, we use a stacked encoder network from the Transformer architecture \cite{vaswani2017attention} to generate normalized importance scores, which are then utilized to create the video summary via the 0/1 knapsack algorithm, following prior work \cite{vslstm}.
\subsection{Preprocessing}
\label{ssec:blip2}
We utilize BLIP-2 \cite{li2023blip}, a state-of-the-art image captioning model, to generate 3 captions per frame before filtering them. 
BLIP-2 has three main components to its architecture---a frozen image encoder, a trainable querying Transformer, and a frozen language model. 
The querying Transformer has image-text matching loss that is able to filter Image-Grounded Text Generation, leading to effective caption generation. BLIP-2 applies the beam search method for generating captions. 
In our model, we use the nucleus sampling method that BLIP-2 also supports for computational efficiency over beam search. 
We embed the generated captions using the SBERT \cite{sbert} model. 
We use the mean pooling method of SBERT, which allows us to correlate each word-level token to a single value in the sentence embedding. 

\subsection{Sentence Filtering and Condensing}
\label{ssec:sentextract}
Humans have the ability to provide a concise summary of a (long) video they watched using natural language. 
These natural language descriptions of the video offer a condensed (semantic) alternative representation of the video compared to a summary using multiple images sampled per second, as all previous literature has attempted. 
Inspired by the human natural language description approach to video summarization, the first part of the task is to filter and condense a sequence of images into one sentence representing one second-long clip. To achieve this salient description, we leverage cosine similarity and the k-means method to find the most ``centered'' description. 
\begin{table*}
    \centering
    \begin{tabular}{lccc|c c|c c|c c}
        \toprule
        \multirow{2}{*}{\textbf{Method}} & \multirow{2}{*}{\textbf{Feature}} & \multirow{2}{*}{\textbf{SumMe}} & \multirow{2}{*}{\textbf{TVSum}} & \multicolumn{2}{c|}{\makecell{\textbf{Performance} \\ \textbf{Difference (\%)}}} & \multicolumn{2}{c|}{\makecell{\textbf{Input Size} \\ \textbf{(MB)}}} & \multicolumn{2}{c}{\makecell{\textbf{Input Size} \\ \textbf{Difference (\%)}}}\\
        & & & & \underline{SumMe} & \underline{TVSum} & \underline{SumMe} & \underline{TVSum} & \underline{SumMe} & \underline{TVSum} \\
        \midrule
        VSLSTM \cite{vslstm} & GoogLeNet & 37.2 & 53.7 & -34.85 & -20.44 & 30.05 & 96.30 & 0 & 0 \\
        ddpLSTM \cite{vslstm} & GoogLeNet & 38.6 & 54.7 & -32.40 & -18.96 & 30.05 & 96.30 & 0 & 0 \\
        SUM-GAN$_{sup}$ \cite{mahasseni2017unsupervised} & GoogLeNet & 41.7 & 56.3 & -26.97 & -16.59 & 30.05 & 96.30 & 0 & 0 \\
        SUM-DeepLab \cite{rochan2018video} & GoogLeNet & 48.8 & 58.4 & -14.54 & -13.48 & 30.05 & 96.30 & 0 & 0 \\
        DSNet \cite{zhu2020dsnet} & GoogLeNet & 50.2 & 62.1 & -12.08 & -8.00 & 30.05 & 96.30 & 0 & 0 \\
        MAVS \cite{feng2018extractive} & GoogLeNet & 43.1 & \textbf{67.5} & -24.52 & 0 & 30.05 & 96.30 & 0 & 0 \\
        CLIP-It$_{L_c}$ \cite{narasimhan2021clip} & CLIP & 49.1 & 60.2 & -14.01 & -10.81 & 16.36 & 51.6 & -45.56 & -46.42 \\
        CLIP-It \cite{narasimhan2021clip} & CLIP & 54.2 & 66.3 & -5.08 & -1.78 & 16.36 & 51.6 & -45.56 & -46.42 \\
        PGL-SUM \cite{apostolidis2021combining} & GoogLeNet & \textbf{57.1} & 61.0 & 0 & -9.63 & 30.05 & 96.30 & 0 & 0 \\
        (Ours) No Input & Tensor & 38.8 & 43.9 & -32.05 & -34.96 & 12.28 & 38.73 & -59.13 & -59.78 \\
        (Ours) LMVS$_{unf}$ & SBERT & 44.4 & 60.1 & -22.24 & -10.96 & 24.54 & 77.37 & -18.34 & -19.66 \\
        (Ours) LMVS & SBERT & 45.8 & 60.5 & -19.79 & -10.37 & 12.28 & 38.73 & -59.13 & -59.78 \\
        \bottomrule
    \end{tabular}
    \caption{Comparison of Supervised F-Scores across models for two datasets. Input size is the size of all feature representations in MBs used as inputs for the models. 
    Performance Difference and Input Size Difference columns show the percentage difference between the state-of-the-art model and other models. 
    LMVS$_{unf}$ represents our model without the filter and condensing method.}
    \label{tab:comp}
\end{table*}
From the 6 fps sample frames, we generate 3 captions per frame using BLIP-2 \cite{li2023blip}. 
With $n$ frames of sequence ($f_1$, $f_2$ ... $f_n$), we would have captions ($c_{1,1}$, $c_{1,2}$, $c_{1,3}$ ... $c_{n,3}$) where each caption is represented as $c_{frame\#,caption\#}$. 
The captions are filtered using k-means and cosine similarity to one representative caption per second.

For the first filtration and condensation method, we employ an approach based on cosine similarity metrics. Using SBERT \cite{sbert}, we create an embedding ($e_{1,1}$, $e_{1,2}$, $e_{1,3}$ ... $e_{n,3}$) for each of the captions. 
Then, we create a pair-wise cosine similarity matrix between the embeddings for each second of the video: \[M_{x,y} = \frac{\Sigma_{i=1}^{d}E_{x,i}E_{y,i}}{\sqrt{\Sigma_{i=1}^{d}E_{x,i}}\sqrt{\Sigma_{i=1}^{d}E_{y,i}}}.\]
Each index of the matrix would show the similarity between the two embeddings $x$ and $y$. 
To filter out and condense the captions to one representative caption, we select one caption that is the most similar to all other captions, $s_i$, where $i$ represents each second of the video: \[s_i = \operatorname*{argmax}_x \Sigma_{y}M_{x,y}.\]

The second method utilizes the naive k-means method \cite{kmeans} to find the centroid of the embedding vectors. 
With the assumption that there exist outliers when sampling lesser-confident outputs from BLIP-2, we partition the embedding vectors into two clusters and find the centroids $C$ = \{$c_1$, $c_2$\}. 
Using $p$, the fixed probability of selecting a point independent of the preceding ones \cite{kmeans}, $C$ can be represented as \[C = \operatorname*{argmin}_c\Sigma_{i=1}^{k}\int_{S_i} |e - c_i|^2 dp(e).\]
Then, using euclidean distance to C, we filter the embedding that is most similar to the selected centroids: \[s_i = \operatorname*{argmax}_e \Sigma_{i=1}^{k} \sqrt{\Sigma_{j=1}^{d}(e_i - c_{i,j})^2}.\]

\subsection{Encoder-Driven Score Generation}
\label{ssec:encstack}

Using a Transformer encoder \cite{vaswani2017attention}, LMVS allows the model to complete self-attention amongst the filtered sentence embeddings. 
This allows the model to generate an importance score based on all sentence descriptions of the videos. 
By allowing a maximum sequence length of 768 captions, we are effectively allowing the model to run its inference on the entire video for all videos in the datasets. 
The maximum sequence length was heuristically chosen as it showed the best performance for our unfiltered and filtered methods. 
Since the longest video in the TVSum dataset is 648 seconds and in the SumMe dataset it is 389 seconds, LMVS can process the entire video for both. 
The ability to process the entire video has empirically improved our model's performance.

The stack of encoders predicts the score for the given subsequence of a video. We employ the normalized score loss to predict the accurate scores: \[L_s = \frac{1}{N}\Sigma_{i\in N}(s_i - \hat{s_i})^2,\] where $N$ is the total number of sentences describing the videos we input into the model, $\hat{s}$ denotes the predicted score for the corresponding sentence, and $s$ represents the normalized score of the ground-truth provided by the dataset. 
Finally, the model output is the normalized importance score of the frame. 
By processing the encoder output through a feed-forward network to produce a scalar, the model assigns importance scores to individual video subsections.

\section{Experiments}
\label{sec:dataset}
To evaluate our model, we use SumMe \cite{summe} and TVSum \cite{tvsum} datasets. We perform two types of experiments for each model; unfiltered top captions sampled at 2 fps and filtered and condensed captions at 1 caption per second.

\subsection{SumMe}
\label{ssec:summe}
The SumMe dataset \cite{summe} consists of 25 videos covering holidays, events and sports. Because they are raw or minimally edited user videos, they have a high compressibility compared to already edited videos. The length of the videos ranges from about 1 to 6 minutes for SumMe. The dataset provides frame-level scores produced by 15 to 18 human subjects. We normalize the scores by subtracting the minimum and dividing by the range for each video.

\subsection{TVSum}
\label{ssec:tvsum}
The TVSum dataset \cite{tvsum}  contains 50 videos representing various genres such as news, how to’s, user generated content. The videos belong to 10 categories and the dataset provides five videos per category. The videos in the dataset vary in length between 2 to 10 minutes. TVSum provides shot-level importance scores annotated via crowdsourcing, which we normalize using the same method applied to SumMe.

\subsection{Experimental Settings}
\label{ssec:exset}
We generate the keyshot-based summary to be less than 15\% of the original video duration, following previous protocols in \cite{vslstm}. Also following the definition of precision (P), overlapped duration of prediction and ground-truth divided by the duration of prediction, and recall (R), overlapped duration of prediction and ground-truth divided by the duration of ground-truth, in \cite{vslstm}, we calculate the F-score; F = (2 x P x R)/(P + R) x 100\%. To ensure an accurate assessment, we employ a random selection approach. We create five random splits and test each split by training on the rest, reporting the average F-score across all five splits. 

\section{Results}
\label{sec:results}
The results are presented in Table \ref{tab:comp}; the input size denotes the size in MB of each method's input feature representations.
We compare our model's performance and input sizes against those of current methods with similar setups. 
When tested on the SumMe dataset, our model exhibited a 19.79\% F-score drop compared to PGL-SUM \cite{apostolidis2021combining}, while consuming 59.13\% less data. 
On the TVSum dataset, the model showed a 10.37\% performance loss relative to MAVS \cite{feng2018extractive}, while using 59.78\% less input data. 
This result is especially notable when considering insights from \cite{ke22efficient}, which showed the correlation between F-score and downsampling rate. 
With the supervised DSNet \cite{zhu2020dsnet} model, the authors showed that input decreasing by 60\% (5 to 2 fps) leads to 22.23\% and 12.07\% drops in F-scores for SumMe and TVSum, respectively \cite{ke22efficient}. 
With a decrease of almost exactly the same input size, our language-only model achieves better results than results using vision modality with decreased inputs. 
Moreover, when matched against CLIP-It$_{L_c}$ \cite{narasimhan2021clip}, which has a single cross-entropy loss function, our model outperforms on the TVSum dataset and has a 6.7\% lower F-score on SumMe.

One limitation of the present work on this initial language-only approach to video summarization is with respect to a detailed exploration of the extensively researched loss functions that have proven to enhance performance in vision-based methods. 
Preliminary experiments showed that the cross-entropy loss function in \cite{vslstm, mahasseni2017unsupervised, rochan2018video, zhu2020dsnet,narasimhan2021clip}, reconstruction loss in \cite{mahasseni2017unsupervised, rochan2018video,narasimhan2021clip}, and diversity loss in \cite{mahasseni2017unsupervised, rochan2018video, narasimhan2021clip} failed to increase performance in our model. 
When we compare our results to the performance of CLIP-It$_{L_c}$, we observe that the language model does perform well under similar conditions. This analysis indicates that our language modality approach is effective and offers potential for enhancement through the integration of multiple, language-appropriate loss functions.

We additionally explore how the signals from the sentence embeddings had a significant impact on performance. 
We achieved this by running a baseline (No Input) with nonsensical input---all-ones tensor for all inputs. 
With this setup, the model was trained solely on signals from the ground-truth annotations.
The results showed that the ground-truth annotations were important signals, but the sentence embeddings had a major influence on the performance. 
Furthermore, the difference in performance is smaller for the state-of-the-art method of SumMe and No Input compared to that of TVSum. 
This finding suggests that the SumMe dataset, containing only half as many videos as TVSum, may not be the best representative dataset for the task of video summarization.

\begin{figure}[t]
  \centering
  \includegraphics[width=8.5cm]{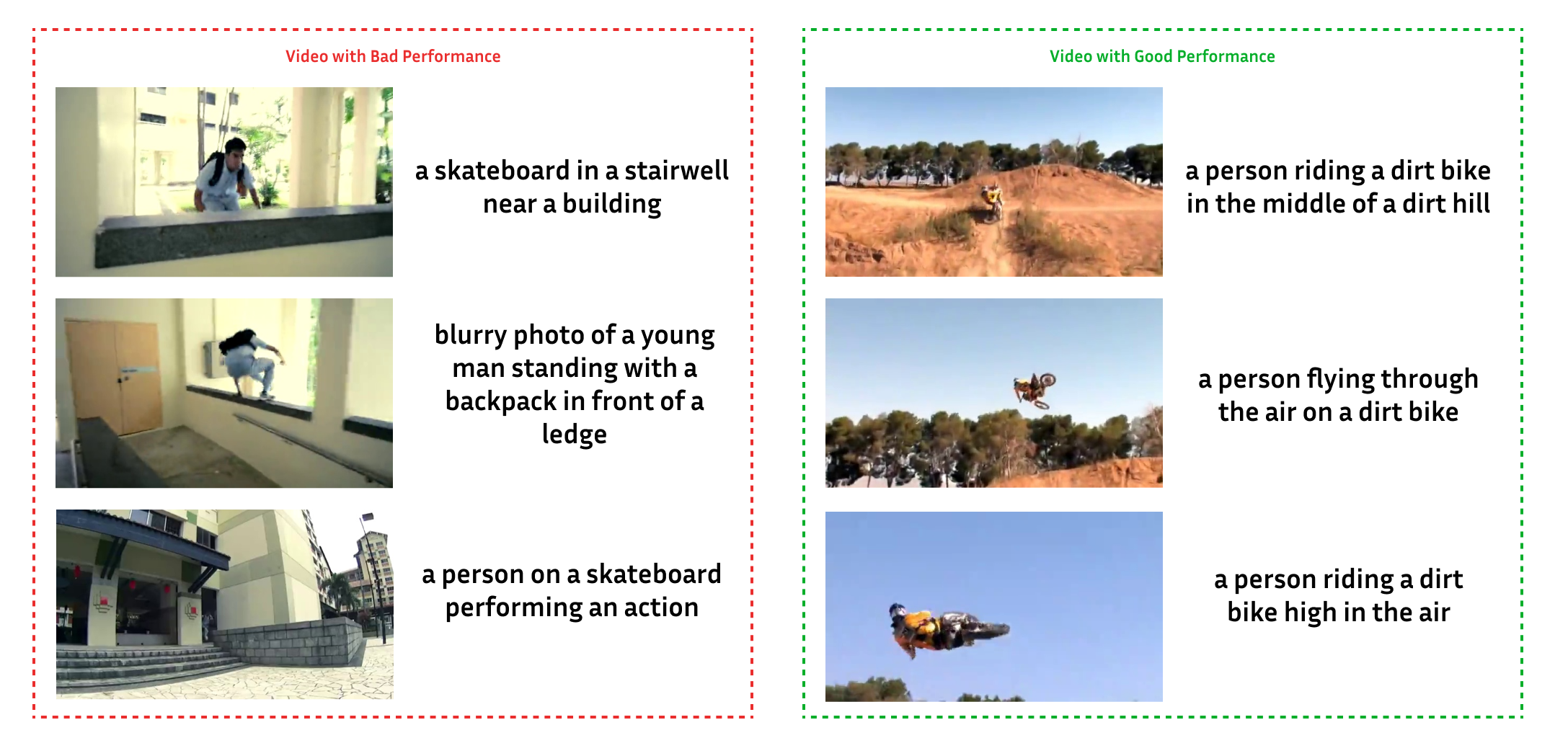}
  \caption{Caption samples corresponding to keyframes}
  \label{fig:outex}
\end{figure}

Lastly, we analyzed our model's weaknesses by examining its performance in 10 different categories of the TVSum dataset. The model consistently struggled with Parkour (PK) and Parade (PR) but excelled in the Attempting Bike Tricks (BT) category, scoring 12 percentage points higher than the PK and PR category on the F-score average. Errors often stemmed from captioning model bias leading to mislabeled captions. Figure \ref{fig:outex} shows examples of wrongly generated captions leading to poor performance. In the PK video on the left where our model drastically underperformed, the captions incorrectly mention a skateboard 42 times, despite the absence of a skateboard in the footage. This suggests that BLIP-2 model has bias, such as suddenly introducing the skateboard object in a video without skateboards, and this could have compromised our model's performance in interpreting the video context. Conversely, the BT video on the right benefited from clear, accurate captions such as ``flying through the air", which likely provided strong signals for the model.

\section{Conclusion}
\label{sec:conclusion}
In this paper, we introduce a novel language-based approach, the LMVS model, to improve data efficiency in video summarization by transitioning from traditional vision only to language modality. Our model either outperforms or achieves comparable results to the best-performing model with a single loss function, both using considerably less data compared to the state-of-the-art. By employing language modality, our model efficiently runs inference on the full video for TVSum and SumMe datasets, generates detailed textual summaries, and enhances interpretability due to the ease with which humans understand natural language. Finally, our experimental results demonstrate the potential of a language-only approach for lightweight video processing. In the future, we will explore techniques to leverage the extractive text summary to augment the video summarization datasets.

\label{sec:refs}

\bibliographystyle{IEEEbib}
\bibliography{refs}

\end{document}